\documentclass[sigconf]{acmart}
\AtBeginDocument{%
  \providecommand\BibTeX{{%
    \normalfont B\kern-0.5em{\scshape i\kern-0.25em b}\kern-0.8em\TeX}}}

\setcopyright{acmcopyright}
\copyrightyear{2024}
\acmYear{2024}
\setcopyright{acmlicensed}\acmConference[HRI '24]{Proceedings of the 2024 ACM/IEEE International Conference on Human-Robot Interaction}{March 11--14, 2024}{Boulder, CO, USA}
\acmBooktitle{Proceedings of the 2024 ACM/IEEE International Conference on Human-Robot Interaction (HRI '24), March 11--14, 2024, Boulder, CO, USA}
\acmDOI{10.1145/3610977.3635003}
\acmISBN{979-8-4007-0322-5/24/03}

\usepackage{caption}
\usepackage[belowskip=-2pt, aboveskip=3pt]{caption}
\usepackage[belowskip=-1pt, aboveskip=2pt]{subcaption}
\usepackage[justification=centering]{subcaption}
\usepackage{amsthm}
\usepackage{amsmath}
\usepackage{mathtools, nccmath}
\usepackage{bbm}
\usepackage[ruled,norelsize,linesnumbered,noend]{algorithm2e}
\SetKwInput{KwInit}{Initialize}
\usepackage{algpseudocode}
\usepackage{wrapfig}
\usepackage[export]{adjustbox}
\usepackage[capitalise]{cleveref}

\usepackage{tikz}
\usetikzlibrary{calc}
\usetikzlibrary{graphs}
\usetikzlibrary{trees}
\usetikzlibrary{shapes.geometric}
\usetikzlibrary{positioning}

\usepackage{placeins}

\usepackage{soul}

\begin{document}

\title{Workspace Optimization Techniques to Improve Prediction of Human Motion During Human-Robot Collaboration}

\author{Yi-Shiuan Tung}
\email{yi-shiuan.tung@colorado.edu}
\affiliation{%
  \institution{University of Colorado Boulder}
  \streetaddress{1 Th{\o}rv{\"a}ld Circle}
  \city{Boulder}
  \country{USA}}

\author{Matthew B. Luebbers}
\email{matthew.luebbers@colorado.edu}
\affiliation{%
  \institution{University of Colorado Boulder}
  \streetaddress{1 Th{\o}rv{\"a}ld Circle}
  \city{Boulder}
  \country{USA}}

\author{Alessandro Roncone}
\email{alessandro.roncone@colorado.edu}
\affiliation{%
  \institution{University of Colorado Boulder, Lab0 Inc.}
  \streetaddress{1 Th{\o}rv{\"a}ld Circle}
  \city{Boulder}
  \country{USA}}

\author{Bradley Hayes}
\email{bradley.hayes@colorado.edu}
\affiliation{%
  \institution{University of Colorado Boulder}
  \streetaddress{1 Th{\o}rv{\"a}ld Circle}
  \city{Boulder}
  \country{USA}}

\renewcommand{\shortauthors}{Yi-Shiuan Tung, Matthew B. Luebbers, Alessandro Roncone, \& Bradley Hayes}

\begin{abstract}
  Understanding human intentions is critical for safe and effective human-robot collaboration. While state of the art methods for human goal prediction utilize learned models to account for the uncertainty of human motion data, that data is inherently stochastic and high variance, hindering those models' utility for interactions requiring coordination, including safety-critical or close-proximity tasks. Our key insight is that robot teammates can deliberately configure shared workspaces prior to interaction in order to reduce the variance in human motion, realizing classifier-agnostic improvements in goal prediction. In this work, we present an algorithmic approach for a robot to arrange physical objects and project ``virtual obstacles'' using augmented reality in shared human-robot workspaces, optimizing for human legibility over a given set of tasks. We compare our approach against other workspace arrangement strategies using two human-subjects studies, one in a virtual 2D navigation domain and the other in a live tabletop manipulation domain involving a robotic manipulator arm. We evaluate the accuracy of human motion prediction models learned from each condition, demonstrating that our workspace optimization technique with virtual obstacles leads to higher robot prediction accuracy using less training data.
\end{abstract}





\maketitle

\begin{figure}[ht]
    \centering
    \includegraphics[width=\linewidth]{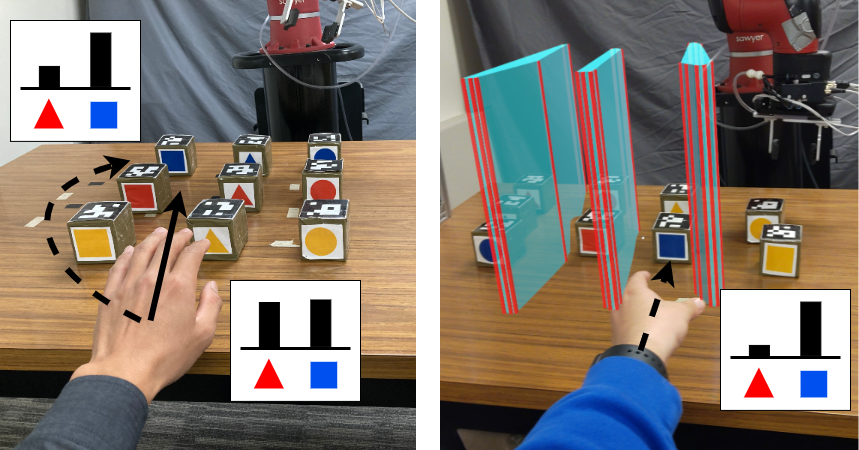}
    \caption{Workspace configuration affects the robot's ability to correctly predict the human's goal -- the blue square cube. \textbf{Left}: The legible path (dotted) requires the human to take a circuitous route while the natural path (solid) is not legible. \textbf{Right}: Our approach generates a workspace configuration by arranging physical objects and projecting ``virtual obstacles'' in AR (cyan and red barriers), in order to induce naturally legible paths from the human.}
    \label{fig:problem}
\end{figure}

\section{Introduction}


\begin{figure*}[!ht]
    \centering\small
    \includegraphics[width=\textwidth]{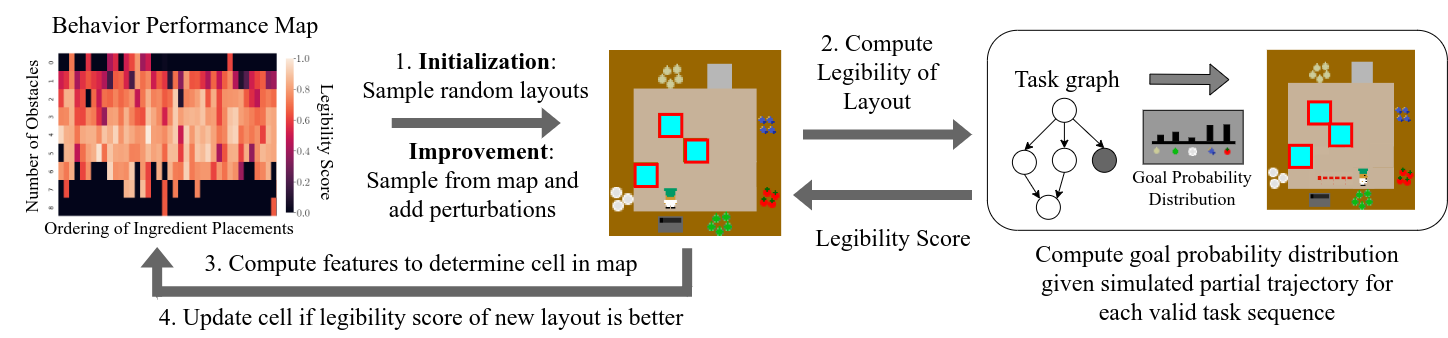}
    \caption{Our approach for generating workspace configurations that enable accurate human goal predictions. \textbf{(1)} In the initialization phase, we sample random environment layouts to populate the behavior performance map, which stores diverse and high performing solutions. This is followed by the improvement phase where we sample directly from the map and add perturbations to test whether the legibility is improved. \textbf{(2)} In both phases, we compute the legibility of the sampled layout by computing the probability of predicting the correct goal at each stage of the task execution. \textbf{(3)} We compute the features of the sampled layout to determine its location in the map. \textbf{(4)} The map is updated if the legibility score of the sampled layout is better than the existing one.}\label{fig:system-diagram}
\end{figure*}

In human-robot collaborative tasks, shared mental models between agents enable the awareness and joint understanding required for effective teamwork \cite{tabrez2020survey}. With no shared notion of the task to be completed, the inherent stochasticity and opacity of human decision-making makes robot planning difficult \cite{sheridan2016human}. To this end, prior research efforts have focused on developing robots that can predict human behavior \cite{rudenko2020human, lasota2017predictor, mainprice2016predict}, generating motion plans to safely interact in a shared environment \cite{pellegrinelli2016human, gupta2022intention}. However, these methods are limited by the quality of robot predictions of a human collaborator's intention and resultant behavior. With inaccurate human models or unexpected human behavior diverging from past experiences, the robot may produce unsafe interactions \cite{markkula2022accurate}, especially in safety-critical or close-proximity settings \cite{vasic2013safety}.

To address the inherent challenges of accurately predicting human motion early in a demonstrated trajectory, our key insight is that robots can take an active role in structuring the environment to reduce the variance of human motion caused by dense and overlapping task spaces, thereby improving the performance of human behavior models.
%
%
%
%
In this work (see \cref{fig:problem}), we introduce an algorithmic approach for a robot to configure a shared human-robot workspace prior to interaction in order to improve a robot's ability to predict the human collaborator's goals during task execution. As detailed in \cref{fig:system-diagram}, we present an objective function that scores potential workspace configurations in terms of how legible the actions of a human teammate are likely to be when performing a task in that environment. We use the mathematical formulation of legibility from \cite{draganlegibility}, which computes the probability of successfully predicting an agent's goal given an observation of a snippet of its trajectory. Our approach finds workspace configurations that maximize legibility over the valid goals at each stage of task execution.

Each candidate workspace configuration combines a potential arrangement of physical objects and projection of ``virtual obstacles'' in augmented reality (AR) in the environment. 
While the arrangement of physical objects can be achieved prior to the shared task via a simple composition of robotic pick and place actions, the addition of augmented reality-based virtual obstacles is particularly effective in imposing explicit constraints on the possible motions of the human, without requiring additional physical changes to the environment.
Note that our approach does not restrict the human to a particular goal and retains their ability to decide on the goal they're reaching towards. Humans also have the flexibility to alter their decisions midway through a task, since goal prediction is performed at every time step.

We efficiently explore the space of workspace configurations using a quality diversity (QD) algorithm called Multi-dimensional Archive of Phenotypic Elites, or MAP-Elites \cite{mouret2015mapelites}. Instead of finding a single optimal solution, MAP-Elites produces a map of performant solutions along dimensions of a feature space chosen by the designer. MAP-Elites enables efficient and extensive exploration of complex search spaces, leading to higher quality solutions as compared to other search algorithms \cite{cully2015robots, nilsson2021policy}.



We empirically demonstrate that our workspace optimization approach improves the accuracy of human goal prediction models, showing results for both a Bayesian predictor with learned cost functions using Maximum Entropy Inverse Reinforcement Learning (MaxEnt IRL) and a time series multivariate Gaussian model. We collect human motion data in an online 2D navigation experiment and a real world tabletop experiment involving a robotic manipulator, comparing our approach to other workspace arrangement strategies. In summary, we present two primary contributions: 1) an algorithm for optimizing the placement of physical and virtual objects in a shared human-robot workspace for maximizing human legibility, and 2) an evaluation of that algorithm via two human subjects experiments, showing that it influences human behavior in ways that improve a robot collaborator's prediction model.




\section{Related Work}
\label{sec:related work}

\emph{Planning using Human Motion Models.}
In human-robot collaboration, the robot needs to predict human motion in order to coordinate its actions with those of the human. Prior work has developed motion planning algorithms that models human motion and generates robot plans to safely interact with humans in tabletop \cite{lasota2015analyzing} and navigation \cite{david2020avoidance, pfeiffer2016predicting} settings. To account for the uncertainty in human motion prediction, prior work has used partially observable Markov decision processes (POMDPs) to determine optimal actions for a robot given a probabilistic belief over the human's intended goals \cite{pellegrinelli2016human,gupta2022intention}. These methods rely on the human motion model to achieve safe interactions, but human motion is inherently highly variable as humans can always move unexpectedly. The difficulty of the problem can be seen by the variety of approaches taken toward accurate human motion prediction: Gaussian models \cite{9300047, perez2015fast}, dynamic movement primitives \cite{8968192, widmann2018dmp}, latent representation learning \cite{butepage2018anticipating, Martinez_2017_CVPR}, and imitation learning \cite{mainprice2016predict, fahad2018pedestrians}, among various other methods \cite{rudenko2020human}. Our work takes a different approach and addresses a fundamental challenge faced by all human motion prediction models; we reduce the uncertainty inherent in modeling the intentions of human collaborators by pushing them towards legible behavior via environment design. Our work improves human motion model predictions by increasing environmental structure to reduce uncertainties, facilitating more fluent human-robot interactions. 


\emph{Environment Design in Robotics:}
Prior work has also explored designing or modifying environments in order to better achieve agent goals \cite{tung2022robotickitting,Tung2023VAMHRI}. \citet{zhang2009general} proposed a framework for designing environments that optimize an agent's reward, and \citet{keren2017equi} extended it for stochastic transitions. \citet{kulkarni2020designing} generate interpretable robot behaviors by modifying the environment. These works show the potential advantages of a robot using environment design to improve its task performance and interpretability. There has also been work on modifying the environment for collaborative teaming. \citet{ijcai2023p611} optimizes warehouse layouts for multi-robot coordination, and \citet{bansal2020supportive} explores the idea of robots moving objects to reduce the likelihood of future collisions in a tabletop task. Our technique differs from prior work in that the robot modifies its environment with the explicit goal of improving its ability to predict human behavior for fluent collaboration with human teammates.

Finding the optimal environment by simply iterating through all possible workspace configurations quickly becomes intractable as the number of objects or possible states increases. To address this, others have used quality diversity (QD) algorithms to generate diverse environments for evaluating the performance of shared autonomy algorithms \cite{fontaine2022evaluating} and explore diverse coordination behaviors as a result of environment design \cite{fontaine2021importance}. Inspired by the success of QD algorithms in finding diverse solutions in large search spaces \cite{cully2015robots, nilsson2021policy}, we use MAP-Elites to search for environments that best elicit legible human behavior. Our work extends QD approaches to generate interaction scenarios that influence human behavior and address the robot's limitations when collaborating with humans.

	

\section{Legible Workspace Generation}
\label{sec:methods}

In this section, we describe our approach for modifying the shared human-robot workspace to maximize legibility and enable more accurate human goal predictions (summarized in Fig. \ref{fig:system-diagram}).


\subsection{Legibility Score} \label{sec: legibility-metric}
To evaluate the legibility of a workspace configuration, we consider the probability distribution of predicting that the human is approaching goal $G$ given an observed trajectory from start state $S$ to intermediate point $Q$. We use the formulation developed by \cite{draganlegibility} shown in Equation \ref{eqn: prob-goal}.

\begin{equation}\small
    \Pr(G | \mathcal{\xi}_{S \rightarrow Q}) \propto \frac{exp(-C(\mathcal{\xi}_{S \rightarrow Q}) - C(\mathcal{\xi}^*_{Q \rightarrow G}))}{exp(-C(\mathcal{\xi}^*_{S \rightarrow G}))}
    \label{eqn: prob-goal}
\end{equation}

The optimal human trajectory from point $X$ to point $Y$ with respect to cost function $C$ is denoted by $\mathcal{\xi}^*_{X \rightarrow Y}$. Equation \ref{eqn: prob-goal} evaluates how cost efficient (with respect to $C$) going to goal $G$ is from start state $S$ given the observed partial trajectory $\mathcal{\xi}_{S \rightarrow Q}$ relative to the most efficient trajectory $\mathcal{\xi}^*_{S \rightarrow G}$.

Let $\mathcal{G}$ be the set of valid goals at the current time step. We develop a legibility score (Eqn. \ref{eqn: legibility-metric}) for use in our optimization objective that, for every valid goal at a given time step in the task execution, maximizes the margin of prediction between the human's chosen goal $G_{true} \in \mathcal{G}$ and all other valid goals. If the predicted goal is not $G_{true}$, the score is penalized by a fixed cost $c$ multiplied by the length of the sampled human trajectory $|\mathcal{\xi}_{S \rightarrow Q}|$. Otherwise, the score is the difference of the two highest probabilities (computed by the $margin$ function). The notation $G_{(i)}$ denotes the $i$-th index of a sorted list of length $n$ that represents the goal probabilities ordered from least to most likely given the observed trajectory $\xi_{S\rightarrow Q}$.

\begin{equation}\small
\text{EnvLegibility}(G_{true}) =
    \begin{dcases}
        -c |\mathcal{\xi}_{S \rightarrow Q}|, \text{   if  } \underset{G \in \mathcal{G}}{\arg\max} \Pr(G | \mathcal{\xi}_{S \rightarrow Q}) \neq G_{true} \\
        margin(\mathcal{G}|\mathcal{\xi}_{S \rightarrow Q}) = G_{(n)} - G_{(n-1)}, \text{   otherwise}
    \end{dcases}
    \label{eqn: legibility-metric}
\end{equation}


\subsection{Optimization for Task Legibility} \label{sec: optimization-task}

To generate a workspace configuration with improved legibility of the agent's goals for a task, we maximize the legibility score from Equation \ref{eqn: legibility-metric} for all valid subtask sequences (Eqn. \ref{eqn: legibility-objective}). We use precedence constraints introduced by the structure of the task (i.e., which subtasks are prerequisites for other subtasks) to identify the set of valid subtasks (and thus valid goals $\mathcal{G}$) at any given time step. 


\begin{equation}\small
    \max \sum_{T' \in \text{permutations}(T)} \mathbbm{1}\{\text{valid}(T')\} \times \sum_{t \in T'} \sum_{G \in \mathcal{G}} \text{EnvLegibility}(G)
    \label{eqn: legibility-objective}
\end{equation}

Let $T'$ represent a valid subtask sequence, consisting of all $k$ subtasks of task $T$: $t_1, ..., t_k$, ordered such that for each index $i$, subtask $t_i$ has all precedence constraints satisfied. Each subtask has one or more goals that an agent can reach to complete it. For example, the task ``Set Table'' may have a subtask ``Get/Place plate on place mat'' with multiple satisfying goals (multiple place mats). Additionally, since there are often multiple valid subtask sequences given an observed set of subtasks $t_1, ..., t_i$, $\mathcal{G}$ represents the set of goals corresponding to all uncompleted subtasks with satisfied precedence constraints. For example, after a plate is set, ``Get/Place Fork'' and ``Get/Place Spoon'' may be equally valid as the next subtask to perform. At this point, $\mathcal{G}$ includes potential goals for both fork and spoon placement subtasks. The objective function considers all possible goals that the human might be reaching for at a given stage of task execution and maximizes the probability of correctly predicting the human's chosen goal.

\begin{algorithm}
    \caption{improve\_workspace}
    \label{alg: gradient-descent}
    \KwIn{Workspace configuration $w$, objective score of w $s_w$, objective function $F$, measure function $M$, Solution map S, Solution values V}
    \KwInit{Best configuration $w^* \leftarrow w$, Best score $s^* = s_w$}
    $A = $ sample\_perturbations($w$) \label{ln: perturbations} \\
    \For{$a \in A$} {
        Generate new workspace $w' = $ apply\_perturbation($w$, $a$) \\
        Compute legibility score $s_{w'} = F(w')$ \\
        \If{$s_{w'} > s^*$} { \label{ln: if start}
             Update best workspace $w^* = w'$\\
             Update best score $s^* = s_{w'}$ \label{ln: if end}
        }
        Determine features $\boldsymbol{m} = M(w')$\\
        \If{$S[\boldsymbol{m}] = \emptyset$ or $s_{w'} > V[\boldsymbol{m}]$} { \label{ln: update start}
        Update solution map $S[\boldsymbol{m}] = w'$\\
        Update solution values $V[\boldsymbol{m}] = s_{w'}$ \label{ln: update end}\\
    }
    }

    \If{$w^* \neq w$} { \label{ln: compare w}
        \Return improve\_workspace($w^*$) \label{ln: run gd}
    }
    \Return $w^*$ \label{ln: return best}
\end{algorithm}

\begin{algorithm}
    \caption{Workspace Generation with MAP-Elites}
    \label{alg: map-elites}
    \KwIn{Human Trajectory Generator $G_H$, Objective function $F$, measure function $M$}
    \KwInit{Solution map $S \leftarrow \emptyset$, Solution values $V \leftarrow \emptyset$}

    \For{$i = 1, ... ,N$} {
        \If{$i < N_{init}$} {
            Generate workspace $w =$ random\_workspace() \label{ln: random workspace}
        }
        \Else {
            Sample workspace from map $w =$ random(S) \label{ln: random sol} \\
            Run $w =$ improve\_workspace($w$) \label{ln: gd}
        }
        Determine features $\boldsymbol{m} = M(w)$ \label{ln: features} \\
        Determine objective score $s = F(w)$ \\
        \If{$S[\boldsymbol{m}] = \emptyset$ or $s > V[\boldsymbol{m}]$} {
            Update solution map $S[\boldsymbol{m}] = w$\\
            Update solution values $V[\boldsymbol{m}] = s$ \label{ln: store value}\\
        }
    }    
    \Return S, V
\end{algorithm}


\subsection{Search using Quality Diversity} \label{sec: map-elites}
Iterating through all possible workspace configurations to find the optimal solution is  intractable since the number of possible configurations is exponential in the number of goals, virtual obstacles, and size of the workspace. We use MAP-Elites \cite{mouret2015mapelites} to approximate the optimal solution.

In Algorithm \ref{alg: map-elites}, MAP-Elites maintains a behavior performance map, or solution map $S$, that stores high performing solutions across features or behaviors of interest. To find the most legible workspace (objective function $F$), the designer chooses a set of features or behaviors (computed by measure function $M$) such as the distance between the objects or the number of virtual obstacles. The algorithm would then find the most legible workspace for each possible combination of features found. As input, the algorithm also requires a model $G_H$ that outputs human trajectory given a goal. $G_H$ can be learned from data via inverse optimal control \cite{mainprice2016predict} or approximated via shortest path to goal \cite{draganlegibility}.

MAP-Elites (Alg. \ref{alg: map-elites}) consists of two phases: initialization and improvement. In the initialization phase, we randomly sample workspaces for $N_{init}$ iterations (Line \ref{ln: random workspace}) and store them in the corresponding cell in the solution map by computing the features (Lines \ref{ln: features}-\ref{ln: store value}). In the solution map $S$, the cell associated with the vector of feature values $m$ is denoted $S[m]$. For $N-N_{init}$ iterations, we perform the improvement phase: we first randomly sample from the solution map (Line \ref{ln: random sol}) and then empirically approximate the gradient (Line \ref{ln: gd}) to improve the solution. Following differentiable QD \cite{fontaine2021differentiable}, we use gradient information to speed up search. Unlike \cite{fontaine2021differentiable}, our implementation only estimates gradients of the objective function and not the measure function. Since we empirically approximate the gradient using stochastic sampling (Alg. \ref{alg: gradient-descent}), the environment updates may not always align with the steepest ascent direction. This can lead to "suboptimal" solutions that can, however, contribute to increased diversity.

As detailed in Algorithm \ref{alg: gradient-descent} (the \textit{improve\_workspace} function), Line \ref{ln: perturbations} samples perturbations to the workspace (i.e. changing an item's position, adding or removing a virtual obstacle). For each perturbation, a new workspace configuration $w'$ is generated by applying the perturbation. We keep track of the current best workspace $w^*$ in terms of the objective score (Lines \ref{ln: if start}-\ref{ln: if end}) and update the solution map if a better environment was found for the features $\boldsymbol{m}$ (Lines \ref{ln: update start}-\ref{ln: update end}). If there was an improvement to the workspace, we run Algorithm \ref{alg: gradient-descent} again (Lines \ref{ln: compare w}-\ref{ln: run gd}). Otherwise, a local minima has been found, and we return the best workspace found (Line \ref{ln: return best}).


\section{Evaluation}
We evaluate our approach in two environments: a) single player Overcooked \citep{carroll2019utility} and b) tabletop collaborative pick and place with a Sawyer robot. These experiments validate our approach across navigation and tabletop domains while analyzing two distinct forms of human-borne motion data. To predict the human's goal given a partial trajectory in Overcooked, we use the Bayesian predictor developed by \citet{draganlegibility} (Eqn. \ref{eqn: prob-goal}). For the tabletop experiment, we implement the time series multivariate Gaussian model proposed in \cite{perez2015fast} to predict which cube the human is reaching for. We chose these models as representative examples of two distinct approaches to human motion prediction. Collectively, these experiments demonstrate the broad applicability of our approach across varied environments and tasks, as well as varied human goal classification techniques.

\subsection{Hypotheses} 
\textit{H1:} Environments generated by our approach will enable more accurate predictions of the human goal throughout the task execution compared to baseline environments.\\ 
\textit{H2:} The prediction models will have better accuracy than predictions based on heuristics such as predicting the goal that is closest to the current trajectory.\\ 
\textit{H3:} Prediction models trained in environments generated by our approach will be more data-efficient, requiring fewer examples to reach peak performance levels as compared to those trained in baseline environments.

\subsection{Overcooked Experiment}

\subsubsection{Experimental Setup}

\begin{figure}[htbp]
\captionsetup[subfigure]{aboveskip=2pt,belowskip=1pt}
\centering
    \begin{subfigure}[b]{0.32\linewidth}
        \centering
        \includegraphics[width=\linewidth]{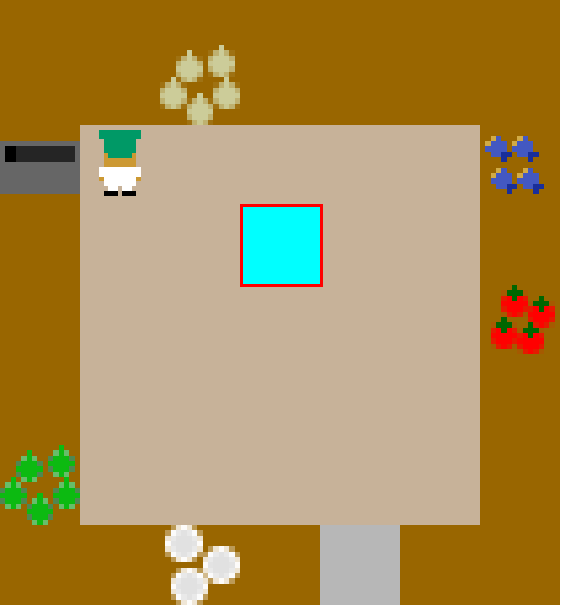}
        \caption{Random\\ \text{}}
        \label{fig:random-layout}
    \end{subfigure}
    \begin{subfigure}[b]{0.32\linewidth}
        \centering
        \includegraphics[width=\linewidth]{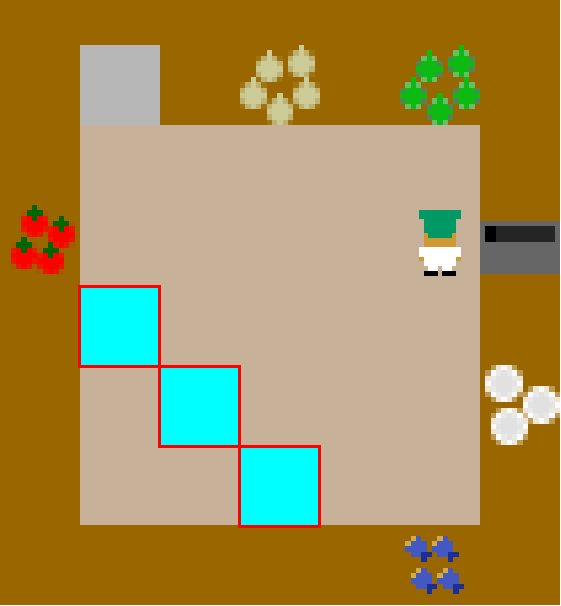}
        \caption{Efficient\\ \text{}}
        \label{fig:efficient-layout}
    \end{subfigure}
    \begin{subfigure}[b]{0.32\linewidth}
        \setcounter{subfigure}{2}
        \centering
        \includegraphics[width=\linewidth]{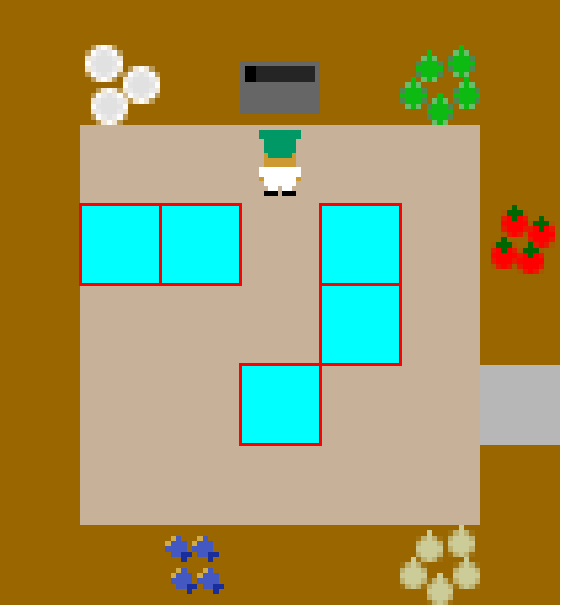}
        \caption{Legible\\ + Efficient}
        \label{fig:legible-efficient-layout}
    \end{subfigure}
    \begin{subfigure}[t]{0.32\linewidth}
        \centering
        \includegraphics[width=\linewidth]{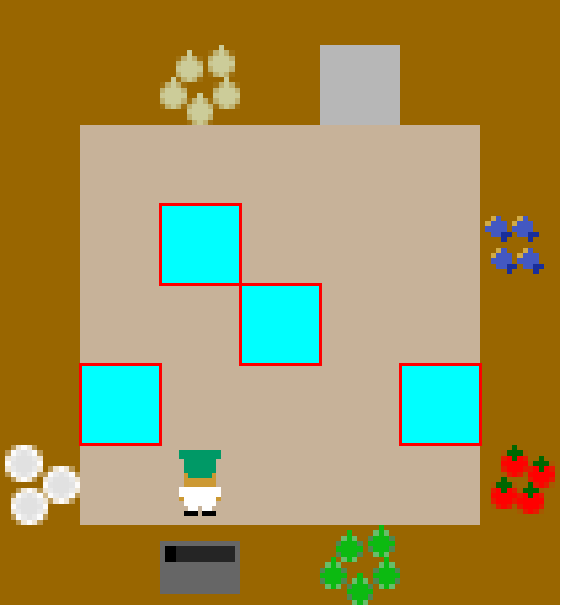}
        \caption{Legible\\ \text{}}
        \label{fig:legible-layout}
    \end{subfigure}
    \begin{subfigure}[t]{0.45\linewidth}
        \setcounter{subfigure}{4}
        \centering
        \includegraphics[width=\linewidth, height=7.3em]{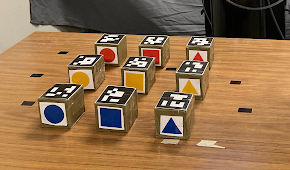}
        \caption{Baseline}
        \label{fig:NoAR-Sorted-layout}
    \end{subfigure}
    \begin{subfigure}[b]{0.45\linewidth}
        \centering
        \includegraphics[width=\linewidth, height=7.3em]{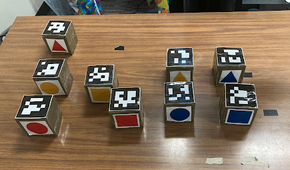}
        \caption{Placement Optimized}
        \label{fig:NoAR-optimized-layout}
    \end{subfigure}
    \begin{subfigure}[b]{0.45\linewidth}
        \setcounter{subfigure}{6}
        \centering
        \includegraphics[width=\linewidth, height=7.3em]{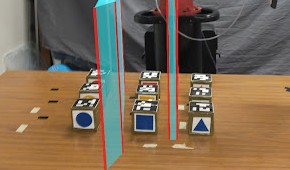}
        \caption{Virtual Obstacle Optimized}
        \label{fig:AR-sorted-layout}
    \end{subfigure}
    \begin{subfigure}[b]{0.45\linewidth}
        \centering
        \includegraphics[width=\linewidth, height=7.3em]{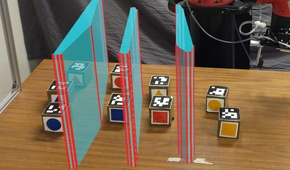}
        \caption{Both Optimized}
        \label{fig:AR-optimized-layout}
    \end{subfigure}
        \begin{subfigure}[b]{0.45\linewidth}
        \centering
        \includegraphics[width=\linewidth]{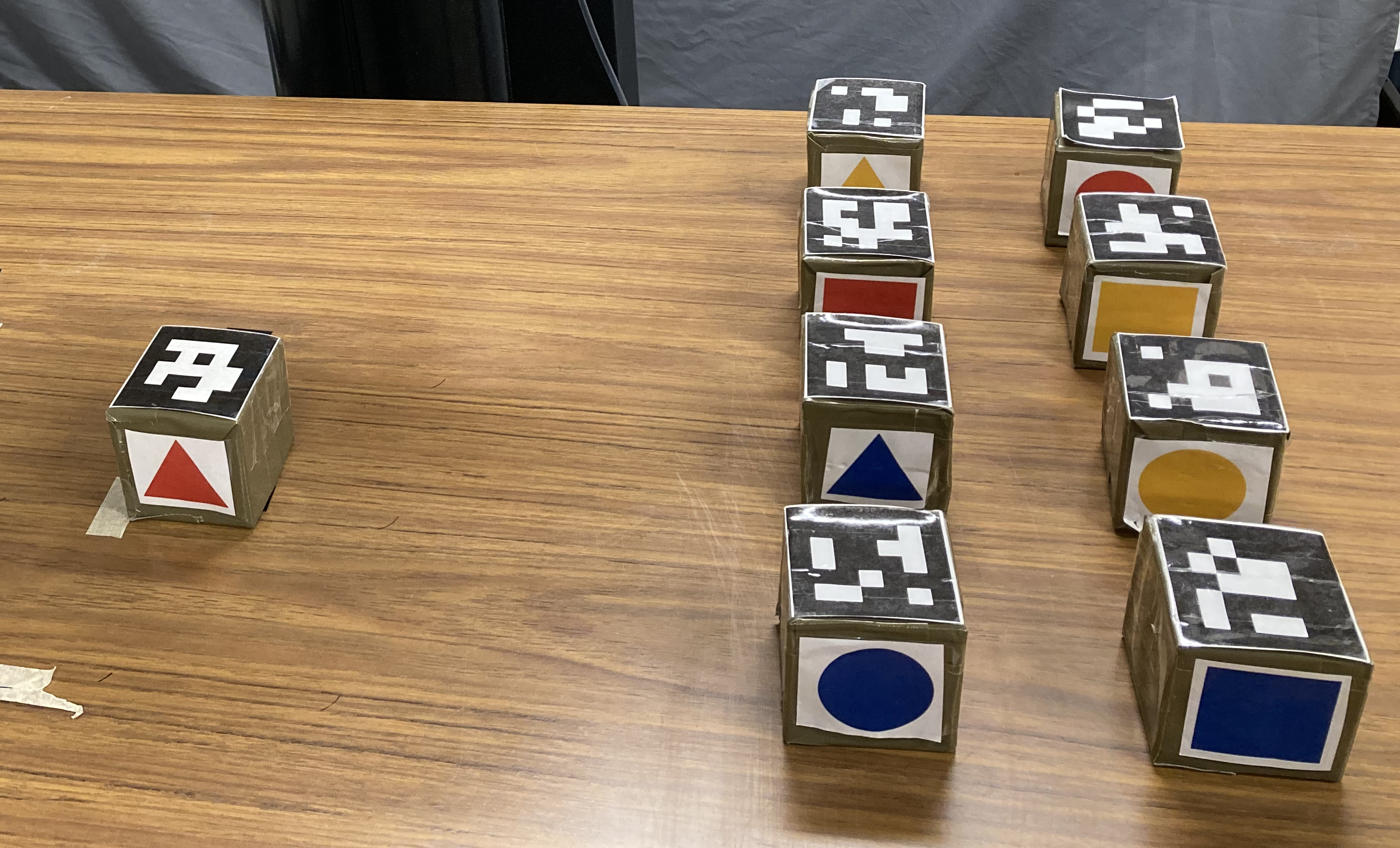}
        \caption{Desired configuration}
        \label{fig:goal config}
    \end{subfigure}
\caption{(a-d) Overcooked layouts and (e-h) initial cube configurations used in our experiments. The environments generated by our approach, (d) \textit{Legible} and (h) \textit{Both Optimized}, optimize object and virtual obstacle (shown in cyan with red edges) placements to elicit legible human motion. (i) The tabletop experiment involves the human and the robot collaboratively placing cubes into the desired configuration -- two columns on the right with a given ordering.}
\label{fig:overcooked-layouts}
\end{figure}

The Overcooked game (Fig. \ref{fig:overcooked-layouts}a-d) requires participants to fetch ingredients, place them in a pot, plate the cooked dish, and deliver the dish to a serving station in a grid world environment. We chose to run the experiment with a single human agent in order to isolate the effects of environment design on the goal predictability of human agents. An investigation on combining robot policy and environment design to influence goal prediction is an interesting direction for future work. We conducted an IRB-approved Amazon Mechanical Turk experiment with 20 participants aged 18 or older and with at least a $95\%$ approval rating. Each participant played in five rounds---one training round followed by four conditions in randomized order. Each round had three soup deliveries with the following ingredients: 2 tomatoes + 1 onion, 2 onions + 1 cabbage, and 3 fish. The soups and the ingredients could be delivered or placed in the pot in any order. At each time step, we recorded the game state and action taken by the participant.

\subsubsection{Goal Prediction Model}

We use the Bayesian formulation (Equation \ref{eqn: prob-goal}) to predict the human goal in the Overcooked experiment. The predicted goal is given by $arg\max_G P(G|\xi)$ where $\xi$ is the observed partial trajectory. We use Maximum Entropy Inverse Reinforcement Learning (MaxEntIRL) \cite{ziebart2008maximum} to estimate the cost function $C_{\theta}$ that is modeled as a linear function of the state features $f_s$, $C_{\theta}(s) = \theta^Tf_s$. We used the position of the player as the features in our experiments. The cost of a trajectory is the sum of the cost of states in the trajectory. MaxEntIRL models the trajectories $\xi$ from the training data as a Boltzmann distribution $p_{\theta}(\xi) = \frac{1}{Z}exp(-C_{\theta}(\xi))$, where $Z$ is the partition function computed via dynamic programming. We optimize the cost function parameters $\theta$ by maximizing the likelihood given training data $\max_{\theta} \sum_{\xi \in D} p(\xi|\theta)$ \cite{ziebart2008maximum}.

\begin{figure*}
    \centering
    \includegraphics[width=\linewidth]{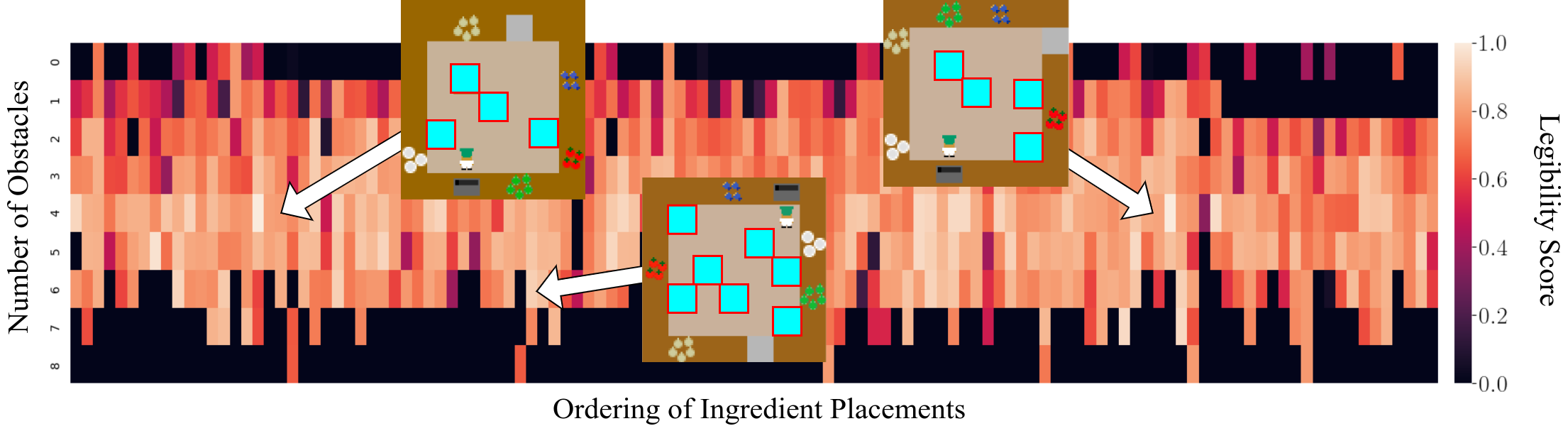}
    \caption{Behavior performance map generated in the Overcooked domain. MAP-Elites is able to explore legible workspace configurations across combinations of the features: number of obstacles and the ordering of ingredient placements, which improves the best solution found in complex search spaces.}
    \label{fig:overcooked-grid}
\end{figure*}

\subsubsection{Conditions}

We compare four conditions (\cref{fig:overcooked-layouts}a-d):\\
\textbf{Random:} We sample 1000 environments where each workstation is randomly placed in the perimeter, and each inner cell has a $20\%$ chance of being assigned a virtual obstacle. This is the same as the initialization step of the MAP-Elites implementation. We then sort the layouts by the legibility score and select the layout with the median score (Fig. \ref{fig:random-layout}).\\
\textbf{Efficient:} We use MAP-Elites to minimize the distance traveled to complete a task (Fig. \ref{fig:efficient-layout}). This condition represents a common objective when designing workspace configurations -- optimizing for task efficiency.\\
\textbf{Legible-Efficient:} The MAP-Elites objective function includes terms for both maximizing legibility and minimizing the distance traveled (Fig. \ref{fig:legible-efficient-layout}).\\
\textbf{Legible:} Our approach -- using MAP-Elites to maximize legibility (Fig. \ref{fig:legible-layout}).

\subsubsection{Workspace Generation}

The workstations of the Overcooked game consist of the cooking pot, dish, serving, and four ingredient stations: onions, tomatoes, cabbages, and fish. For multiple agents to efficiently collaborate, an agent has to predict the ingredients the other agents are picking up so that efforts aren't duplicated. Therefore, for the objective function and evaluation, we only consider trajectories starting from the pot or serving stations and ending at the dish or ingredient stations. We allow the algorithm to move the workstations in the perimeter which is representative of reorganizing objects. The algorithm can also add or remove virtual obstacles onto any space within the perimeter. We add a constraint that the workstations have to be at least two cells away so that trajectories will be longer than a single step.

All conditions, except the Random condition, are generated using MAP-Elites with different objective functions. The features are the number of obstacles and the ordering of workstation placements from top to bottom and left to right. Figure \ref{fig:overcooked-grid} shows the behavior performance map generated by MAP-Elites. After 1000 iterations of the improvement phase, we use the solution with the highest legibility score in the map.

\textbf{\textit{1) Initialization.}} The workspace is randomly sampled by finding non-overlapping placements for each workstation in the perimeter. For each cell not in the perimeter, there is a $20\%$ chance the cell is a virtual obstacle. We verify that the sampled workspace has valid paths between the workstations before placing it in the behavior performance map. This phase is run for $1000$ iterations.

\textbf{\textit{2) Improvement.}} The available perturbations are moving a station to an empty counter, swapping positions with another station, and adding or removing virtual obstacles.

\subsection{Tabletop Experiment}

\subsubsection{Experimental Setup}
We conducted an IRB-approved human subjects study with 12 participants (8 male, 4 female), with an age range of 19 to 31 years old (M = 24.42, SD = 3.37), recruited from a university campus to collect human hand trajectory data in different tabletop workspace setups. We collected 8 trajectories per participant for each of the four conditions, presented in a randomized, counterbalanced order. The study involves the human and an autonomously operating Sawyer 7 DoF manipulator robot working collaboratively to place cubes into a desired configuration (Fig. \ref{fig:goal config}). To prevent the robot from picking up the same cube, the participant is asked to first reach for a cube while the robot maintains a probability distribution over the possible cubes the human is reaching for in real time. Once the robot is sufficiently confident of the human's goal, the robot will select its own cube to pick up and move to grasp it. The human and robot team continues to pick a cube each until the task is completed. The precedence constraints are set such that the first column of the desired configuration must be completed before the second column can start. This experiment evaluates homogeneous teams where the human and robot are able to perform all subtasks. Our approach generalizes to heterogeneous teams, in which case we modify $T$ from Equation \ref{eqn: legibility-objective} to only include subtasks the human can perform.

\subsubsection{Goal Prediction Model} \label{sec: tabletop prediction model}

\sloppy We use a time series multivariate Gaussian model \cite{perez2015fast} to predict human goals given the human hand trajectory collected from the tabletop experiment. For a training data set $D$, we use dynamic time warping (DTW) \cite{muller2007dynamic} to align the trajectories so they all have length $K$. For each time step $k \in K$ and goal $G$, we train a multivariate Gaussian with mean $\mu_G[k] = \frac{1}{|D|} \sum_{i=1}^{|D|}{f_i[k]}$ and covariance $\Sigma_G[k] = \frac{1}{|D|-1} \sum_{i=1}^{|D|}{(f_i[k] - \mu[k])(f_i[k]-\mu[k])^T}$ where $f_i[k]$ are the feature values of the $i$th trajectory at time step $k$. At prediction time, the probability of a goal $G$ given a partial trajectory $\xi$ is $P(G|\xi) \propto \prod_{k=1}^K [\mathcal{N}(\mu_G[k], \Sigma_G[k])]^{\frac{1}{K}}$ \cite{perez2015fast}. The goal with the highest probability is the predicted goal. We use two features, the $x$ and $y$ positions of the hand, captured via an Intel RealSense RGB camera and a real-time hand tracking algorithm \cite{mediapipe}.

\begin{figure}
    \centering
    \begin{subfigure}[t]{\linewidth}
    \centering
    \scalebox{0.7}{
    \begin{tikzpicture}
        \node [label=right:{Random},fill={rgb,255:red,153; green,204; blue,0}, rounded corners=1pt] (node1) {};
        \node [label=right:{Efficient},fill={rgb,255:red,55; green,126; blue,184}, rounded corners=1pt] (node2) at ([xshift=2cm]node1.east){};  
        \node [label=right:{Legible-Efficient},fill={rgb,255:red,152; green,78; blue,163}, rounded corners=1pt] (node3) at ([xshift=2cm]node2.east){};
        \node [label=right:{Legible},fill={rgb,255:red,255; green,127; blue,0}, rounded corners=1pt] (node4) at ([xshift=3cm]node3.east){};
        \draw[thick, rounded corners=3pt, gray] ($(node1.north west)+(-0.25,0.2)$) rectangle ($(node4.south east)+(1.4,-0.2)$);
    \end{tikzpicture}
    }
    \end{subfigure}
    \begin{subfigure}[t]{0.9\linewidth}
        \centering
        \includegraphics[width=\linewidth]{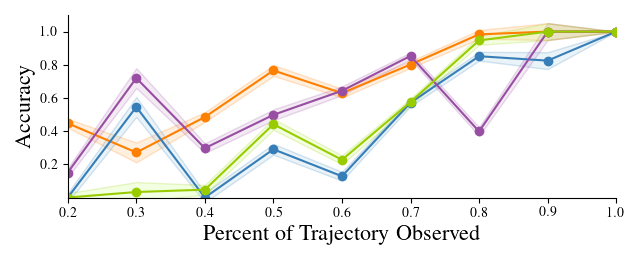}
        \caption{MaxEnt IRL cost function}
        \label{fig:accuracy-irl}
    \end{subfigure}
    \begin{subfigure}[b]{\linewidth}
        \centering
        \includegraphics[width=\linewidth]{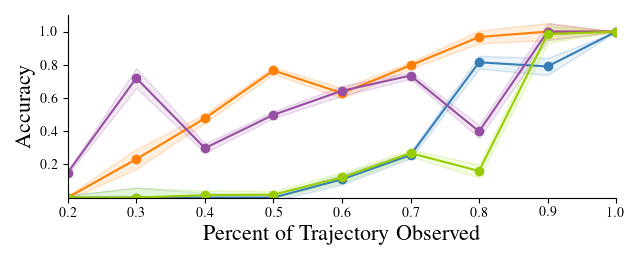}
        \caption{Shortest distance heuristic}
        \label{fig:accuracy-shortest-path}
    \end{subfigure}
    \caption{Overcooked experiment results: (a) Mean accuracy of Bayesian predictor using cost function learned from MaxEnt IRL. Compared to the shortest distance heuristic shown in (b), MaxEnt IRL improved accuracy the most when less than $60\%$ of the trajectory has been observed. The width of the shaded area is the Tukey's Q critical value, so that overlaps indicate no statistical significance. The \textit{Legible} condition elicits significantly higher accuracy for most cases.
    }
    \label{fig:overcooked-accuracy}
\end{figure}
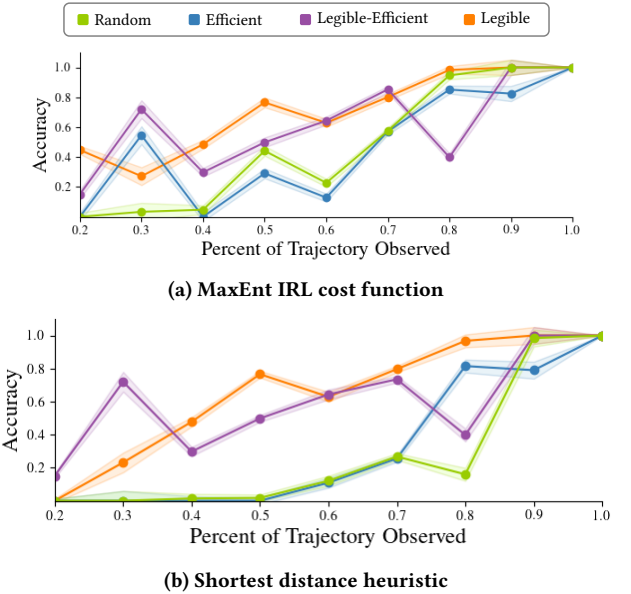

\begin{figure*}
    \captionsetup[subfigure]{aboveskip=-1pt,belowskip=-1pt}
    \centering
    \begin{subfigure}[b]{\textwidth}
    \centering
    \scalebox{0.7}{
    \begin{tikzpicture}       
        \node [label=right:{Baseline},fill={rgb,255:red,0; green,201; blue,128}, rounded corners=1pt] (node1) {};
        \node [label=right:{Placement Optimized},fill={rgb,255:red,0; green,0; blue,128}, rounded corners=1pt] (node2) at ([xshift=1.7cm]node1.east){};  
        \node [label=right:{Virtual Obstacle Optimized},fill={rgb,255:red,255; green,0; blue,255}, rounded corners=1pt] (node3) at ([xshift=3.4cm]node2.east){};
        \node [label=right:{Both Optimized},fill={rgb,255:red,255; green,191; blue,0}, rounded corners=1pt] (node4) at ([xshift=4.2cm]node3.east){};
        \draw[thick, rounded corners=3pt, gray] ($(node1.north west)+(-0.25,0.2)$) rectangle ($(node4.south east)+(2.4,-0.2)$);
    \end{tikzpicture}
    }
    \end{subfigure}
    \begin{subfigure}[b]{0.39\textwidth}
        \centering
        \includegraphics[width=\textwidth]{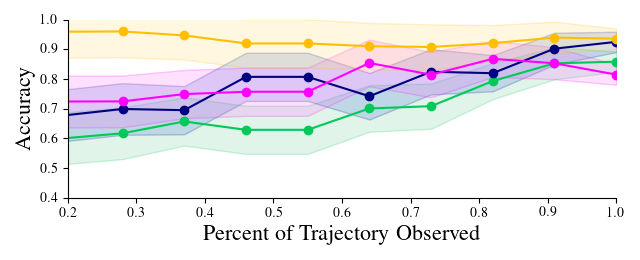}
        \caption{Time series multivariate Gaussian}
        \label{fig:accuracy-lm}
    \end{subfigure}
    \begin{subfigure}[b]{0.39\textwidth}
        \centering
        \includegraphics[width=\textwidth]{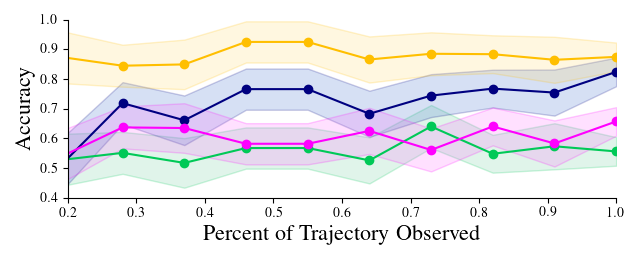}
        \caption{Shortest distance heuristic}
        \label{fig:accuracy-shortest-path-tabletop}
    \end{subfigure}

    \begin{subfigure}[b]{\textwidth}
        \centering
        \includegraphics[width=0.32\textwidth]{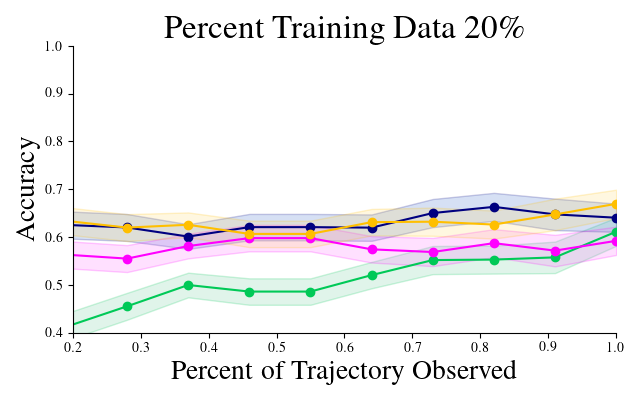}
        \includegraphics[width=0.32\textwidth]{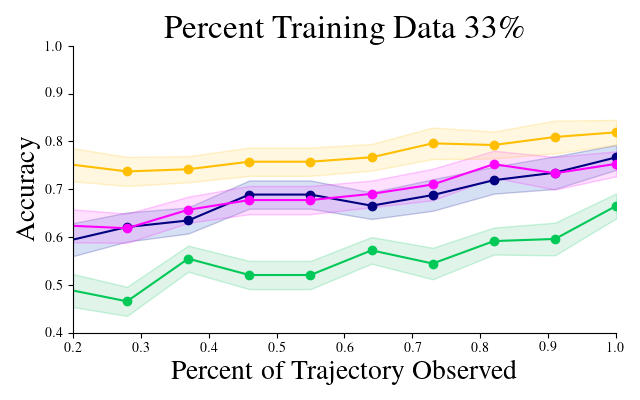}
        \includegraphics[width=0.32\textwidth]{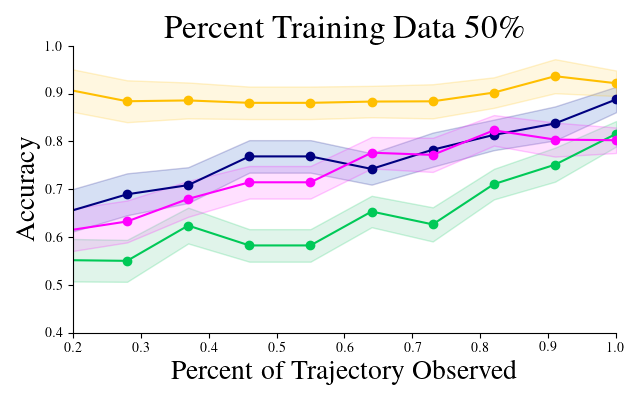}
        \caption{Time series multivariate Gaussian learning curve}
        \label{fig:tabletop-learning}
    \end{subfigure}
    \caption{Tabletop experiment results: (a) Mean accuracy of time series multivariate Gaussian model. Our approach, \textit{Both Optimized}, elicits significantly higher or comparable accuracy compared to the baselines. (b) Mean accuracy using shortest distance heuristic where the closest cube to the current hand position is the predicted goal. (c) Mean accuracy of time series multivariate Gaussian model with varying amounts of training data. \textit{Both Optimized} elicits significantly higher accuracy compared to baselines starting at $33\%$ training data, achieving around $90\%$ accuracy with just $50\%$ of the available training data.}
    \label{fig:tabletop-res}
\end{figure*}
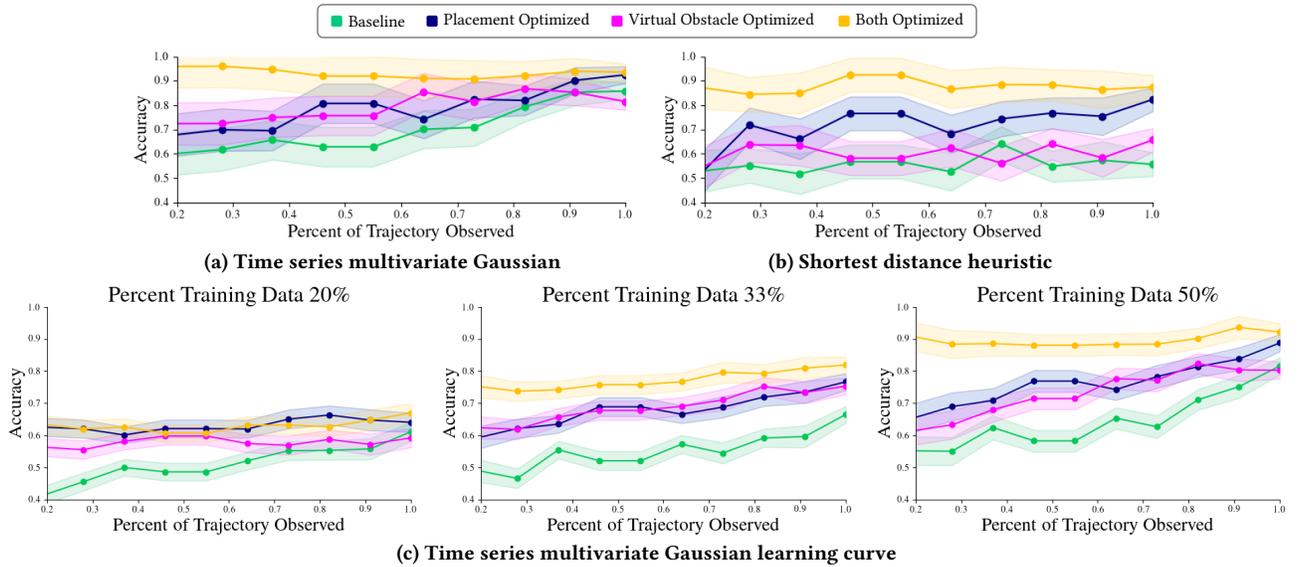

\subsubsection{Conditions}

We compare workspace setups using our optimization against \textbf{Baseline}, where the cubes are initially sorted by their color and shape (Fig. \ref{fig:NoAR-Sorted-layout}). We also perform an ablation study, removing the algorithm's ability to organize the cubes or project virtual obstacles. As such, the remaining conditions are:\\
\textbf{Placement Optimized:} We optimize for legibility by only rearranging the cubes, with no virtual obstacles (Fig. \ref{fig:NoAR-optimized-layout}).\\
\textbf{Virtual Obstacle Optimized:} The cubes are sorted by their color and shape, and we optimize for legibility by projecting virtual obstacles (Fig. \ref{fig:AR-sorted-layout}).\\
\textbf{Both Optimized:} Our approach that optimizes for legibility by rearranging the cubes and projecting virtual obstacles (Fig. \ref{fig:AR-optimized-layout}).

\subsubsection{Workspace Generation}

The feature dimensions for MAP-Elites are the minimum distance between the cubes and the ordering of cubes by the x-axis. In contrast to the layout generation for the Overcooked game, we do not add virtual obstacles in the initialization step. Due to the continuous state space and the difficulty of randomly sampling a useful virtual obstacle, we choose fixed size virtual obstacles and insert them between two randomly selected cubes in the improvement phase. By employing this heuristic, we can effectively explore configurations that result in altered human reaching motions, which are approximated using shortest paths in a visibility graph. We experimented with virtual obstacles of different sizes and found that obstacles measuring 30cm x 1cm were sufficient in inducing distinct reaching motions.

\textbf{\textit{1) Initialization.}} We randomly sample $(x,y)$ positions for the cubes and ensure that they don't overlap. The positions are continuous values uniformly sampled within the permissible boundaries.

\textbf{\textit{2) Improvement.}} The improvement step consists of first changing cube positions and then adding virtual obstacles. We sample new cube positions from a Gaussian centered at the cube with variance $=7$cm. The virtual obstacles are placed between two randomly sampled cubes.

The boundaries of the algorithmically generated virtual obstacles are passed to an AR interface, implemented using a Microsoft HoloLens 2 head-mounted display. The AR interface renders those obstacles directly in the environmental context of the shared workspace as holograms of cyan barriers with red outlines (Fig. \ref{fig:AR-optimized-layout}). These barriers appear to the human as if they are physically located in the environment, and indicate regions of the environment the human should not enter.



\section{Results}

In both experiments, we run a stratified 4-fold cross validation with 3 repeats and different randomization in each repetition. The stratified cross validation preserves the percentage of samples for each class (i.e. the valid goals at each time step) in each fold. Figures \ref{fig:overcooked-accuracy} and \ref{fig:tabletop-res} show the mean goal prediction accuracy as a function of the percentage of total trajectory observed for Overcooked and tabletop experiments respectively. A prediction is defined as correct if the single highest probability value in the probability distribution is the same as the target goal. A trajectory is the human motion when reaching towards a goal: $x$, $y$ positions in the Overcooked grid and hand positions in the $x$-$y$ plane for the tabletop experiment. We use the one-way analysis of variances (ANOVA) and perform post-hoc analysis using Tukey's HSD test for multiple comparisons to test for effects between condition pairs. We plot Tukey's Q critical value as the width of the shaded area in Figures \ref{fig:overcooked-accuracy} and  \ref{fig:tabletop-res} such that overlaps indicate an absence of a statistically significant difference.

\vspace{-0.3em}
\subsection{Overcooked} 
Figure \ref{fig:accuracy-irl} shows the prediction results using the Bayesian predictor. The \textit{Legible} condition elicits significantly higher or comparable accuracy than the best performer among other conditions except when $30\%$ of the trajectory has been observed, supporting hypothesis \textit{H1}. The counterintuitive bump at 30\% is because the optimization penalizes incorrect predictions more for longer trajectories (Eqn. \ref{eqn: legibility-metric}). Thus, it made a trade-off: the accuracy is lower when predicting goals about 30\% into a subtask but higher for longer trajectories. Figure \ref{fig:accuracy-shortest-path} shows the same analysis but we use the shortest Manhattan distance as a heuristic for the cost function. The mean accuracy is lower when only a small percentage of the trajectory has been observed compared to the predictions with learned cost functions, suggesting that IRL enables faster accurate goal predictions which supports hypothesis \textit{H2}.


\subsection{Tabletop Experiment} Figure \ref{fig:accuracy-lm} shows the mean goal prediction accuracy when different percentages of the total trajectory is observed using the Gaussian model. The environment generated by our approach, \textit{Both Optimized}, elicits significantly higher prediction accuracy than baseline environments when less than $50\%$ of the trajectory has been observed. When using the shortest distance heuristic (Fig. \ref{fig:accuracy-shortest-path-tabletop}), where the closest cube to the current hand position is the predicted goal, all conditions elicit lower accuracy compared to the predictions when using the Gaussian models. 
The shortest distance heuristic exhibits poor performance even when the entire trajectory is observed. 
Both hypotheses \textit{H1} and \textit{H2} are supported by these results.

To evaluate the learning curve of the Gaussian model, we perform stratified splits cross validation with 10 repeats. Figure \ref{fig:tabletop-learning} shows the performance as training data increases. \textit{Both Optimized} elicits significantly higher accuracy than the baselines with $33\%$ training data and achieves around $90\%$ accuracy with $50\%$ training data, supporting our hypothesis \textit{H3} that our approach generates environments where the training of prediction models is more data-efficient. 

\begin{figure}
\centering
    \begin{subfigure}[b]{0.46\linewidth}
        \centering
        \includegraphics[width=\linewidth]{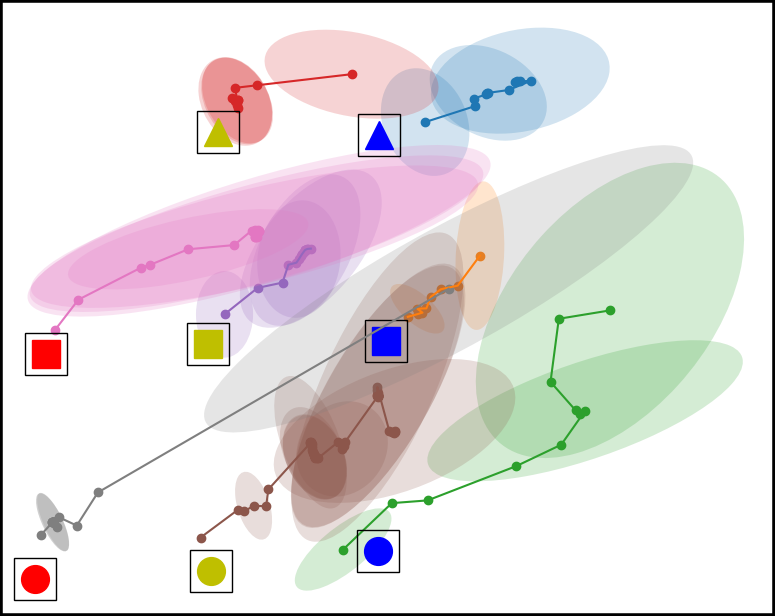}
        \caption{Baseline}
    \end{subfigure}
    \begin{subfigure}[b]{0.46\linewidth}
        \centering
        \includegraphics[width=\linewidth]{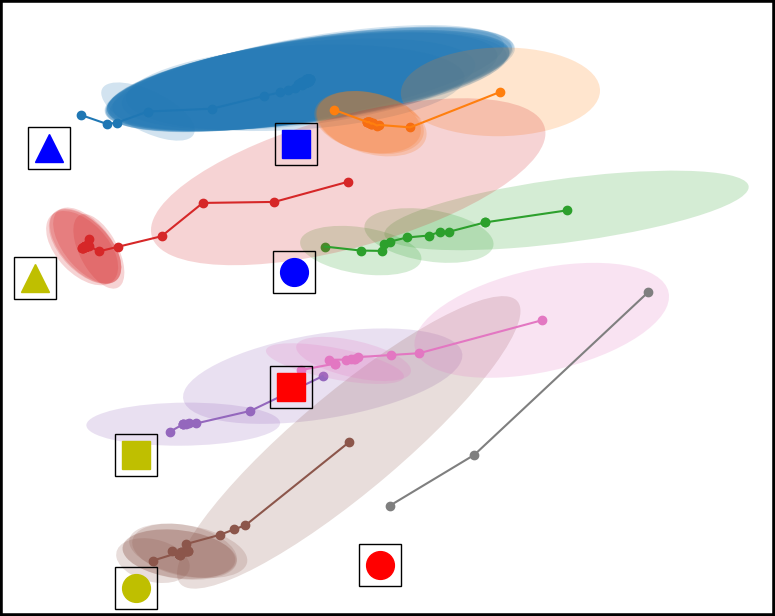}
        \caption{Placement Optimized}
    \end{subfigure}
    \begin{subfigure}[b]{0.46\linewidth}
        \centering
        \includegraphics[width=\linewidth]{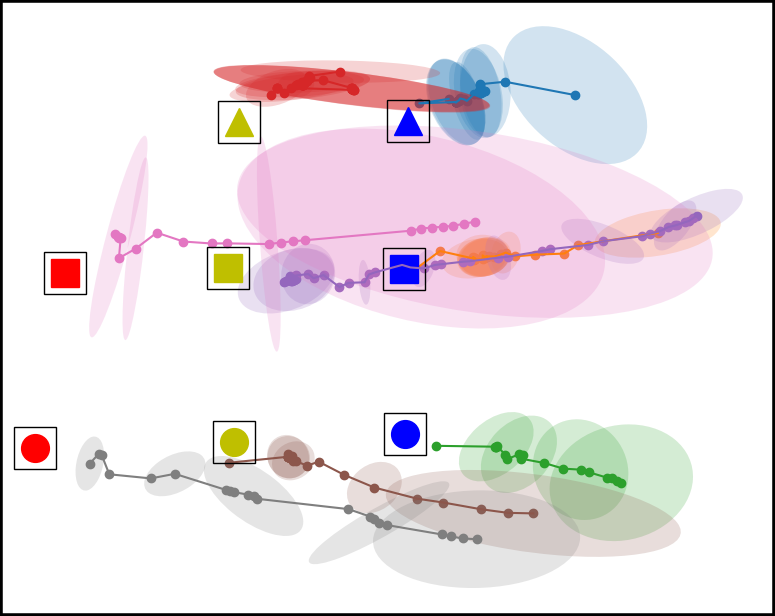}
        \caption{Virtual Obstacle Optimized}
    \end{subfigure}
    \begin{subfigure}[b]{0.46\linewidth}
        \centering
        \includegraphics[width=\linewidth]{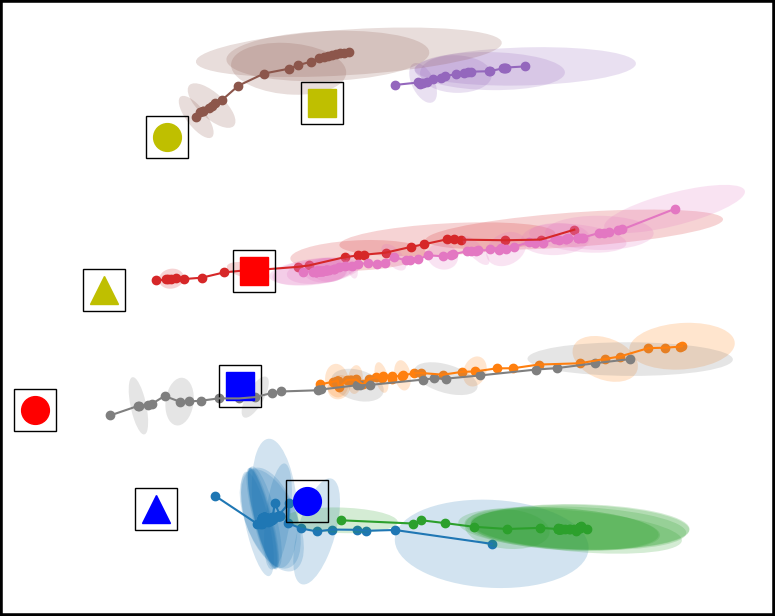}
        \caption{Both Optimized}
    \end{subfigure}
    \caption{Top down view of the workspace plotting the mean and covariance of the time series multivariate Gaussian for each condition. The Gaussian models are fit to the hand trajectory data and predicts the human's goal (section \ref{sec: tabletop prediction model}). The model in the \textit{Both Optimized} condition has less covariance compared to the models trained in the other environment configurations.}
    \label{fig:gaussian-models}
\end{figure}

In Figure \ref{fig:gaussian-models}, we plot the mean and covariance of the learned time series multivariate Gaussian for each condition. Qualitatively, the Gaussian models in the \textit{Both Optimized} condition have less covariance, a measure of uncertainty, compared to the Gaussian models trained in baseline environment configurations. Table \ref{table:gaussians} shows the average determinant and trace of the covariance matrices of the Gaussian models. The determinant is a measure of the magnitude of variation among the variables, and the trace is the sum of the variances of the individual variables but does not consider the correlations between the variables. The Gaussian models trained in the \textit{Both Optimized} layout have lower values for both the determinant and trace compared to the models trained in baseline environments. 

\begin{table}[t]
\caption{The average determinant (det) and trace of the covariance matrices of the multivariate Gaussian models.}
\label{table:gaussians}
\begin{tabular}{ccccc}
    \hline
     & Baseline & Placement & Virtual Obsta- & Both \\
     & & Optimized & cle Optimized & Optimized \\
     \hline
     Det & 9.36e-06 & 8.50e-06 & 6.14e-06 & \textbf{1.09e-06} \\
     Trace & 7.12e-03 & 5.90e-03 & 3.90e-03 & \textbf{2.77e-03}\\
     \hline
\end{tabular}
\end{table}

Overall, our results show that our workspace configuration approach reduces the uncertainty in human motion data and improves the accuracy of goal prediction models in a planar navigation and tabletop manipulation tasks. The models are also more data-efficient in our generated environments, requiring less data to reach their best performance levels as compared to those trained in baseline environments.


\section{Conclusion} \label{sec:discussion}

In this work, we introduce an algorithmic approach for autonomous workspace optimization to improve robot predictions of a human collaborator's goals. By rearranging physical objects in the workspace and projecting AR-based virtual obstacles into the environment prior to interaction, the robot influences the human into more legible behavior during task execution, thereby reducing the uncertainty of their motion. Through dual experiments in 2D navigation and tabletop manipulation, we show that our approach results in more accurate model predictions across two distinct goal inference methods, requiring less data to achieve these correct predictions. Importantly, we demonstrate that environmental adaptations can be discovered and leveraged to compensate for shortfalls of prediction models in otherwise unstructured settings.

Our approach is applicable for domains where the following conditions hold: 1) multiple agents share the same physical space and the agents do not have access to other agents’ controllers or decision making processes, 2) the environment allows physical or virtual configurations, and 3) environment configuration can be performed prior to the interaction. By improving goal prediction, we envision that our framework enhances human-robot teaming across domains such as shared autonomy for assistive manipulation, warehouse stocking, and cooking assistance, among others. Through our results, we demonstrate the generalizability of our overall approach across varied tasks, environments, and human motion prediction models, showing its potential to realize fluency improvements in a number of human-robot collaborative domains.

One limitation is that the improvement of our approach on prediction models is dependent on the task and the environment. If the configurability of the environment is limited, the effect size of our algorithm will likely be small. Future work should explore beyond objective measures of human predictability, investigating the subjective implications of our approach on the human’s experience (i.e. do more complex configurations increase cognitive load for the human?). Lastly it would be valuable to evaluate our approach over extended interactions to determine whether virtual obstacles remain necessary after multiple interactions or if humans naturally adapt to more legible trajectories after “training”.

\begin{acks}
This work was supported by the Office of Naval Research under Grant N00014-22-1-2482 and the Army Research Laboratory under Grant W911NF-21-2-0126.
\end{acks}

\bibliographystyle{ACM-Reference-Format}
\balance
\bibliography{main}  

\clearpage
\appendix

\section{Task Graph}

\begin{figure}[!htb]
\centering
\includegraphics[width=\linewidth]{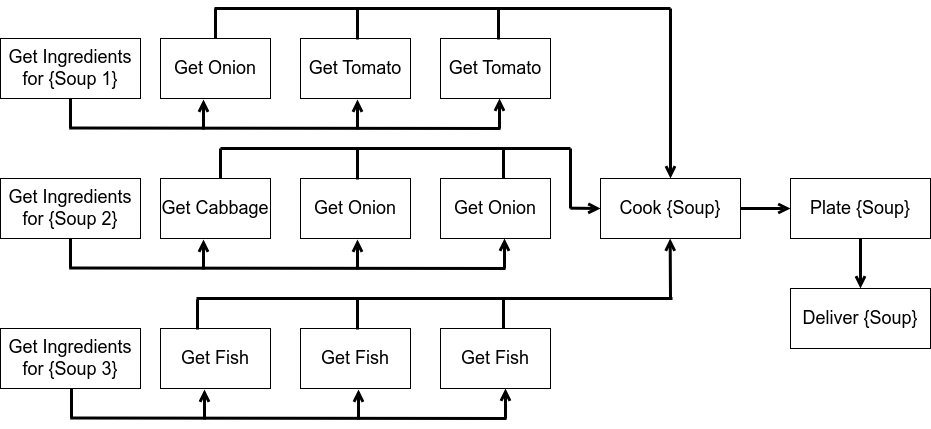}
\caption{Task graph for the Overcooked game environment. Arrows indicate precedence constraints such that if $t_i \rightarrow t_j$ then subtask $t_i$ has to be completed before subtask $t_j$ can begin. A total of 3 soups have to be delivered, and each soup consists of 3 ingredients. The soups and ingredients can be delivered or placed in any order.}
\label{fig:overcooked-task-graph}
\end{figure}

\section{Learned Reward Functions}

\begin{figure}[!htb]
\centering
    \begin{subfigure}[b]{0.24\linewidth}
        \centering
        \includegraphics[width=\linewidth]{figures/simulation/random-layout.drawio.png}
    \end{subfigure}
    \begin{subfigure}[b]{0.24\linewidth}
        \centering
        \includegraphics[width=\linewidth]{figures/simulation/efficient-layout.drawio.png}
    \end{subfigure}
    \begin{subfigure}[b]{0.24\linewidth}
        \centering
        \includegraphics[width=\linewidth]{figures/simulation/legible-efficient-layout.drawio.png}
    \end{subfigure}
    \begin{subfigure}[b]{0.24\linewidth}
        \centering
        \includegraphics[width=\linewidth]{figures/simulation/legible-layout.drawio.png}
    \end{subfigure}
    \begin{subfigure}[b]{0.24\linewidth}
        \centering
        \includegraphics[width=\linewidth]{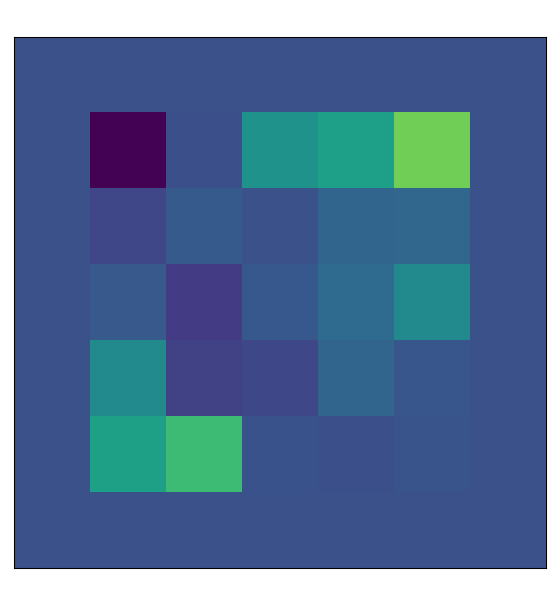}
        \caption{Random \\ \text{}}
    \end{subfigure}
    \begin{subfigure}[b]{0.24\linewidth}
        \centering
        \includegraphics[width=\linewidth]{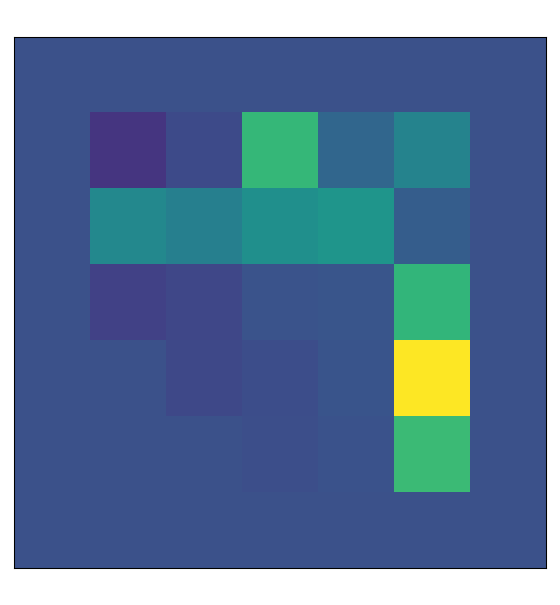}
        \caption{Efficient\\ \text{}}
    \end{subfigure}
    \begin{subfigure}[b]{0.24\linewidth}
        \centering
        \includegraphics[width=\linewidth]{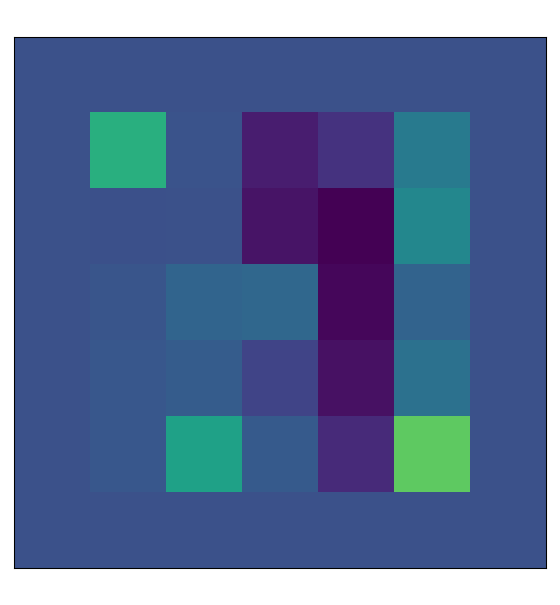}
        \caption{Legible\\ + Efficient}
    \end{subfigure}
    \begin{subfigure}[b]{0.24\linewidth}
        \centering
        \includegraphics[width=\linewidth]{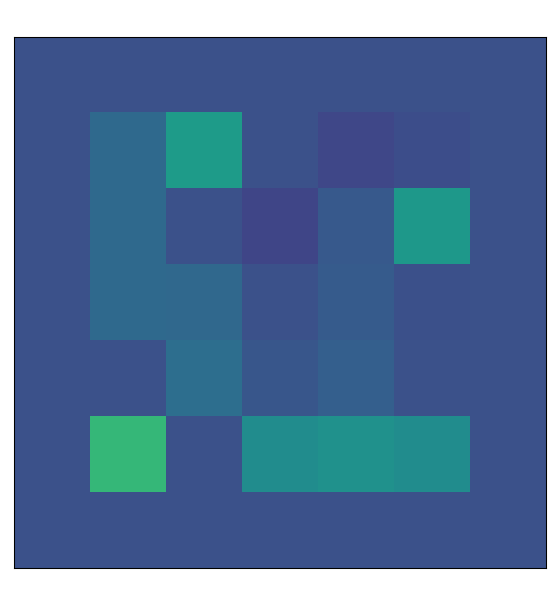}
        \caption{Legible\\ \text{}}
    \end{subfigure}
\caption{Learned reward functions from MaxEnt IRL.}

\label{fig:overcooked-learned-rewards}
\end{figure}

We use MaxEnt IRL to estimate the reward functions for each layout. We use the negative of the reward function as the cost function in Equation \ref{eqn: prob-goal} and predict the target goal of trajectories in the test data set. Figure \ref{fig:overcooked-learned-rewards} shows the learned reward function normalized to values between 0 (dark color) and 1 (bright color) for one of the cross validation splits. As expected, the squares near workstations have higher rewards. We can also observe some preferred paths such as taking the right most lane to go from the pot to onion station in the \textit{Legible-Efficient} condition.




\section{Goal Prediction Results}

\begin{figure}[htbp]
\centering
    \begin{subfigure}[b]{0.45\linewidth}
        \centering
        \includegraphics[width=\textwidth]{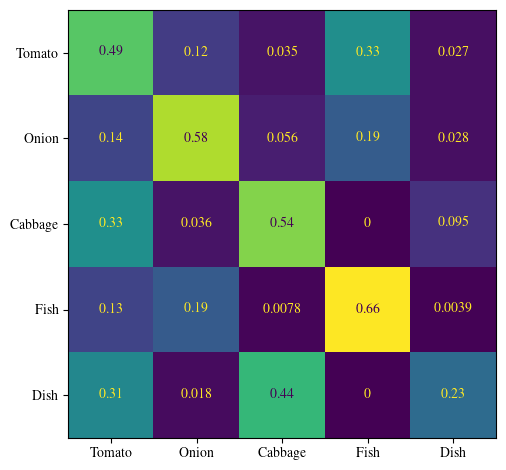}
        \caption{Random}
    \end{subfigure}
    \begin{subfigure}[b]{0.45\linewidth}
        \centering
        \includegraphics[width=\textwidth]{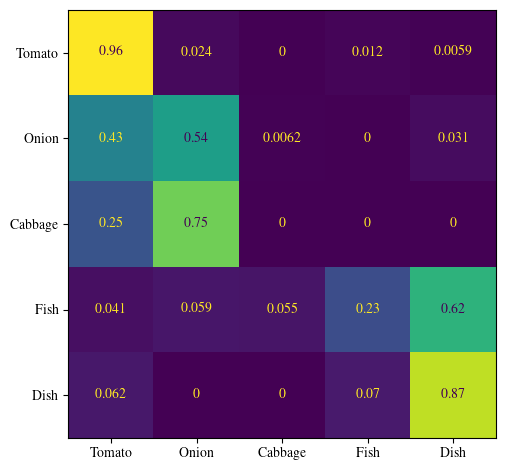}
        \caption{Efficient}
    \end{subfigure}
    \begin{subfigure}[b]{0.45\linewidth}
        \centering
        \includegraphics[width=\textwidth]{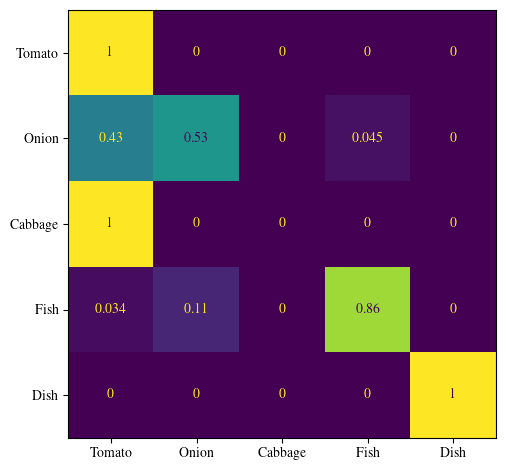}
        \caption{Legible + Efficient}
    \end{subfigure}
    \begin{subfigure}[b]{0.45\linewidth}
        \centering
        \includegraphics[width=\textwidth]{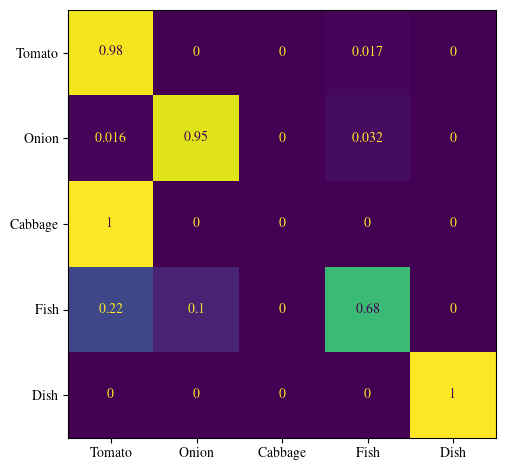}
        \caption{Legible}
    \end{subfigure}
    \caption{Confusion matrix for the Bayesian predictor in the different generated Overcooked layouts.}
    \label{fig:confusion-matrix-overcooked}
\end{figure}

\begin{figure}[htbp]
\centering
    \begin{subfigure}[b]{0.45\linewidth}
        \centering
        \includegraphics[width=\textwidth]{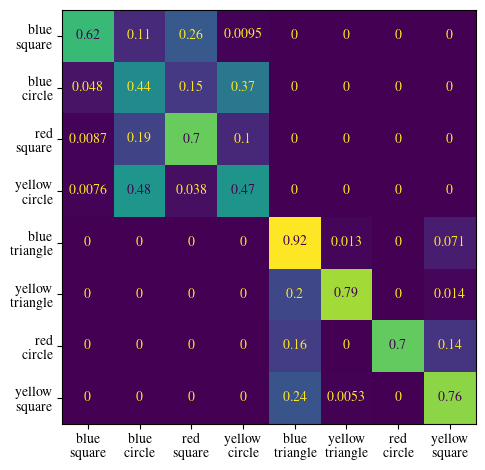}
        \caption{Baseline}
    \end{subfigure}
    \begin{subfigure}[b]{0.45\linewidth}
        \centering
        \includegraphics[width=\textwidth]{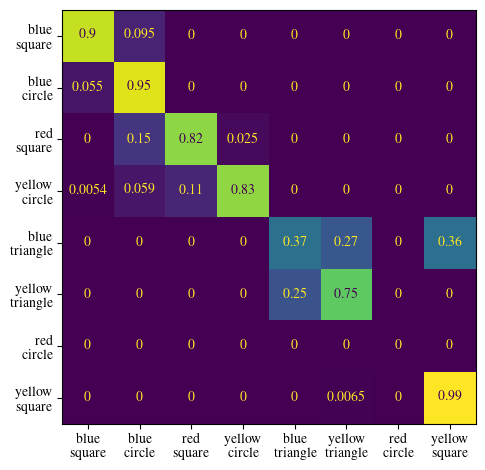}
        \caption{Placement Optimized}
    \end{subfigure}
    \begin{subfigure}[b]{0.45\linewidth}
        \centering
        \includegraphics[width=\textwidth]{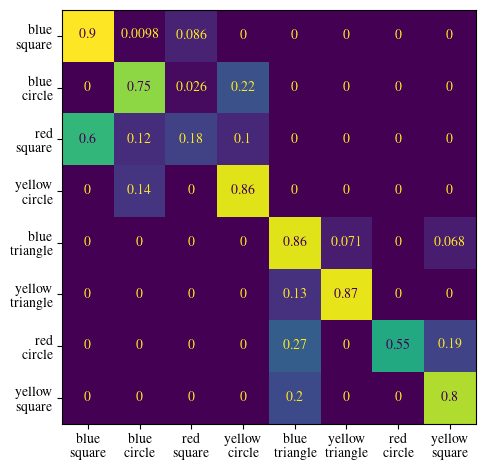}
        \caption{Virtual Obstacle Optimized}
    \end{subfigure}
    \begin{subfigure}[b]{0.45\linewidth}
        \centering
        \includegraphics[width=\textwidth]{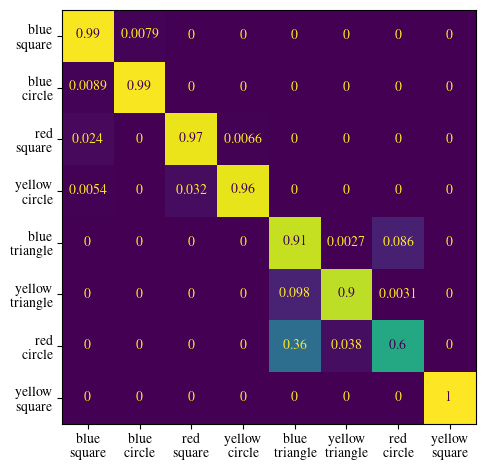}
        \caption{Both Optimized}
    \end{subfigure}
    \caption{Confusion matrix for the time series multivariate Gaussian in the different generated workspace configurations. The x-axis shows the true target, and the y-axis is the predicted target.}
    \label{fig:confusion-matrix-tabletop}
\end{figure}

Figure \ref{fig:confusion-matrix-overcooked} shows the confusion matrix for the goal prediction across the different conditions in Overcooked. The x-axis shows the true target, and the y-axis is the predicted target. The layout generated by our approach, \textit{Legible}, has $100\%$ correct predictions for the target goals tomato and dish. Due to the environment and task complexity, the optimization trades off the ability to predict the goal cabbage for accurate predictions of the other target goals. Cabbage is only used once in the recipes and thus incurs less cost for incorrect predictions. Note that we only include the results when a partial trajectory is observed since the goal prediction is always correct given the entire trajectory.

\begin{table}[htbp]
\centering
\caption{Overcooked Goal Prediction}
\label{table:overcooked-model-res}
\begin{tabular}{ccccc}
    \hline
     & Random & Efficient & Legible + Efficient & Legible \\
     \hline
     F1 & 0.477 & 0.512 & 0.744 & \textbf{0.799} \\
     Precision & 0.564 & 0.648 & 0.792 & \textbf{0.824} \\
     Recall & 0.482 & 0.554 & 0.751 & \textbf{0.812} \\
     \hline
\end{tabular}
\end{table}


\begin{table}[htbp]
\centering
\caption{Tabletop Goal Prediction}
\label{table:tabletop-model-res}
\begin{tabular}{ccccc}
    \hline
     & Baseline & Placement & Virtual Obsta- & Both \\
     & & Optimized & cle Optimized & Optimized \\
     \hline
     F1 & 0.704 & 0.753 & 0.796 & \textbf{0.936} \\
     Precision & 0.751 & 0.770 & 0.803 & \textbf{0.941} \\
     Recall & 0.689 & 0.761 & 0.800 & \textbf{0.937} \\
     \hline
\end{tabular}
\end{table}

Table \ref{table:overcooked-model-res} and \ref{table:tabletop-model-res} show the F1 score, precision, and recall of the goal predictions of the Bayesian predictor in Overcooked and the time series multivariate Gaussian in the tabletop experiments. We account for label imbalance by computing the metrics for each label and computing their average weighted by support. We use the implementation from scikit-learn \cite{scikit-learn}.

\section{Comparisons with other Optimization Methods}

We compared MAP-Elites to Novelty Search with Local Competition (NSLC) \cite{nslc} and Differential Evolution (DE) \cite{Storn1997} for the Overcooked game. We report the average legibility scores across 10 runs with different seeds (larger values are better) in Table \ref{table:optimization-comparisons}.

This result shows that MAP-Elites outperforms NSLC and DE, but the best choice of optimization method is likely conditioned on the task and the environment. In the implementation, we set the hyperparameters such that they have the same number of iterations performing the update step for an individual from the population. MAP-Elites initializes the solution map with 100 random layouts and performs the improvement phase for 100 iterations. NSLC maintains a population size of 20 that evolves for 10 generations. DE performs a maximum of 200 generations over which the entire population is updated.

\begin{table}[htbp]
\centering
\caption{Comparisons of optimization methods on the best solution found (i.e. legibility scores) for different Overcooked environments.}
\label{table:optimization-comparisons}
\begin{tabular}{ccccc}
     \hline
     & Square & Square & Square & Rectangle\\
     & 5x5 & 7x7 & 9x9 & 9x5 \\
     \hline
     DE & -129.28 & -202.52 & -229.22 & -186.31 \\
     & (2.19) & (3.96) & (4.68) & (3.27) \\
     NSLC & -95.44 & -175.43 & -190.54 & -169.63\\
     & (10.73) & (15.92) & (11.89) & (20.39) \\
     MAP-Elites & \textbf{-86.37} & \textbf{-167.88} & \textbf{–173.52} & \textbf{-162.97} \\
     & \textbf{(7.55)} & \textbf{(16.13)} & \textbf{(8.20)} & \textbf{(10.60)}\\
    \hline
\end{tabular}
\end{table}
\FloatBarrier

\end{document}